\documentclass[dvipsnames,format=sigconf]{acmart}
\AtBeginDocument{%
  }

\usepackage{algorithm}
\usepackage{algorithmic}
\usepackage{amsmath}
\usepackage{bm}
\usepackage{cleveref}
\usepackage{subcaption}
\usepackage{xspace}

\usepackage{comment}

\newcommand{\argmax}{\operatornamewithlimits{argmax}} % argmax
\newcommand{\x}{\boldsymbol{x}}

\newcommand{\bb}{\boldsymbol{b}}

\newcommand{\Loss}{\mathcal{L}}
\newcommand{\Data}{\mathcal{D}}
\newcommand{\Archive}{\mathcal{A}}

\newcommand{\MAE}{{\mathrm{MAE}}}
\newcommand{\MedAE}{{\mathrm{MedAE}}}

\usepackage{xcolor}
\newcommand{\markupdraft}[2]{% {#1: {color|display} command}{#2: desired color or text}
    \ifthenelse{\equal{#1}{display}}{#2}{}%                 % display only in draft version
    \ifthenelse{\equal{#1}{color}}{\color{#2}}{}%           % colored only in draft (for \new command)
}
\newcommand{\newcolored}[3][]{{\markupdraft{color}{#2}#3}%  % kept in the final print
    \ifthenelse{\equal{#1}{}}{}{\markupdraft{display}{{\color{yellow!70!black}[#1]}}}} 

\copyrightyear{2026}
\acmYear{2026}
\setcopyright{cc}
\setcctype{by}
\acmConference[GECCO '26]{Genetic and Evolutionary Computation Conference}{July 13--17, 2026}{San Jose, Costa Rica}
\acmBooktitle{Genetic and Evolutionary Computation Conference (GECCO '26), July 13--17, 2026, San Jose, Costa Rica}
\acmDOI{10.1145/3795095.3805087}
\acmISBN{979-8-4007-2487-9/2026/07}

\begin{document}

\title{Diversified Residual Symbolic Regression}

% \authornote{}

\author{Koki Ikeda}
\email{ikeda.k@ic.c.titech.ac.jp}
\orcid{0009-0004-9588-3770}
\affiliation{
    \institution{CyberAgent}
    \city{Shibuya}
    \state{Tokyo}
    \country{Japan}
}
\affiliation{%
    \institution{Institute of Science Tokyo}
    \city{Meguro}
    \state{Tokyo}
    \country{Japan}
    \postcode{152-8550}
}

\author{Masahiro Nomura}
\email{nomura@comp.isct.ac.jp}
\orcid{0000-0002-4945-5984}
\affiliation{%
    \institution{Institute of Science Tokyo}
    \city{Meguro}
    \state{Tokyo}
    \country{Japan}
    \postcode{152-8550}
}

\author{Ryoki Hamano}
\email{hamano_ryoki_xa@cyberagent.co.jp}
\orcid{0000-0002-4425-1683}
\affiliation{%
    \institution{CyberAgent}
    \city{Shibuya}
    \state{Tokyo}
    \country{Japan}
    \postcode{150-0042}
}

\renewcommand{\shortauthors}{Ikeda et al.}

\begin{abstract}
    Symbolic regression (SR) aims to discover explicit mathematical expressions that explain observed data and is widely used in domains where interpretability is essential. Because interpretability requires expressions to reflect meaningful regularities, SR is sensitive to observations that deviate from the dominant relationship. Such irregular observations, or outliers, are common in real-world data and can hinder SR from identifying underlying regularities. Robust regression mitigates this by downweighting observations with large residuals. However, deciding which observations should be treated as outliers is often ambiguous and depends on user interpretation and domain knowledge, a perspective largely overlooked in existing SR studies. This motivates approaches that present multiple candidate expressions, allowing users to examine different residual patterns and choose expressions consistent with their expertise. We propose diversified residual symbolic regression (DRSR), which achieves high predictive accuracy while promoting diversity with respect to residual patterns based on the Quality-Diversity paradigm. DRSR collects multiple expressions that fit the data well but differ in how residuals are distributed, enabling post-search selection aligned with domain knowledge. On a synthetic mixture dataset, DRSR produces more diverse expressions than conventional SR while capturing multiple underlying relationships. On a real-world astronomical dataset, DRSR discovers multiple expressions consistent with known physical relationships.

\end{abstract}

\ccsdesc[500]{Computing methodologies~Machine learning}
\ccsdesc[500]{Computing methodologies~Randomized search}

\keywords{Symbolic regression, Genetic programming, Robustness, Quality Diversity}


\maketitle

% \begin{align}
%     F = ma  \label{eq:equation}
% \end{align}
% % -------------------------------------------------- %
% \subsection{Subsection} \label{ssec:subsection}
% % -------------------------------------------------- %
% % ---------------------------------------- %
% \subsubsection{Subsubsection} \label{sssec:subsubsection}
% % ---------------------------------------- %

% ============================================================ %
\section{Introduction} \label{sec:intro}
% ============================================================ %
Symbolic regression (SR) searches for explicit mathematical expressions that fit observed data.
Unlike black-box models whose internal structure is difficult to inspect, SR produces a closed-form equation that can be read, simplified, and compared with existing knowledge.
SR has been applied across a wide range of domains, including physics \cite{SR-physics:2009}, materials science \cite{SR-materials:2019,SR-materials:2022}, biology \cite{SR-biology:2022,SR-biology:2020}, and economics \cite{SR-economics:2016,SR-economics:2019}.

Genetic programming (GP)~\cite{gp} is a widely used methodology for SR, in which candidate expressions are encoded as tree structures and evolved through selection and variation.
Most GP-based SR systems calculate the fitness of candidates using mean squared error (MSE) or closely related losses.
These losses can be dominated by a small number of large residuals, which has long been a central concern in statistics~\cite{Robust-Rousseeuw:1987}.
In some cases, these residuals are caused by contamination such as measurement errors.
In other cases, they indicate heterogeneity, where the dataset is better viewed as a mixture of distinct regression regimes generated by different underlying models.
In GP-based SR, the search can be affected by such irregular observations, causing it to overlook the expression that best explains the main structure.

To mitigate this effect, existing SR methods for outlier-corrupted data often rely on filtering.
RANSAC-GP~\cite{ransacgp} integrates GP with the RANSAC procedure~\cite{RANSAC:1981, exRANSAC:2007} by repeatedly evolving an expression from a randomly sampled minimal subset and selecting the expression supported by the largest consensus set of inliers.
Sun~et~al.~\cite{Sun:2025} propose a reinforcement learning-based SR framework that uses a dynamic gating module to mask noisy input variables during training, reducing the impact of irrelevant information.
In many applications, however, it is not obvious which observations should be treated as outliers, and the answer can depend on domain knowledge.
As a result, observations that users consider important can be filtered out, and the search may miss expressions that explain those observations well.

From a statistical perspective, the sensitivity of least-squares fitting to outliers has been widely investigated, and robust regression offers a standard response.
The key idea is to reduce the influence of observations with large residuals, so that the fitted model is determined mainly by the majority pattern in the data.
This can be achieved by replacing squared error with a robust estimator \cite{Robust-Huber:1992} or by using high-breakdown estimators that effectively ignore a subset of observations \cite{Robust-Rousseeuw:1984}.
However, minimizing such loss functions often results in a non-convex optimization problem with multiple local minima.
Consequently, different local minima may downweight different observations, implying that the observations treated as outliers can vary across solutions~\cite{robust-local:2004, robust-local:2019}.

More generally, SR research has explored ways to align the returned expression with user goals when data fit alone is not sufficient.
A common approach is multiobjective GP, which simultaneously minimizes prediction error and expression complexity, yielding a Pareto-optimal set that exposes the accuracy and simplicity trade-off.
Another direction injects prior knowledge into SR, for example through shape constraints or physics-informed guidance, to bias the search toward domain-consistent expressions~\cite{Keren:2023, Zhou:2023}.
However, these approaches still evaluate candidates through objectives to be minimized or maximized and therefore do not explicitly separate alternative expressions that differ mainly in which observations they explain well.
Moreover, since prior knowledge is often domain-specific and difficult to transfer, a more general mechanism is needed to preserve diverse high-quality candidates.

In noisy or heterogeneous datasets, multiple expressions can achieve similar predictive error while differing mainly in which observations they explain well.
This is especially problematic when outliers are plausible and their interpretation is uncertain, because standard SR tends to return a single expression that implicitly commits to one residual allocation.
We therefore propose \emph{Diversified Residual Symbolic Regression} (DRSR), a general quality-diversity framework for symbolic regression that aims to produce a diverse set of competitive expressions characterized by residual-based behaviors.
Since residual allocation is not a quantity to minimize or maximize in a single direction, DRSR maintains candidate expressions in a grid archive, as in MAP-Elites~\cite{MAP-Elites:2015}.
While prior QD-based SR studies \cite{QD-SR:2019, QD-SR:2025} mainly use diversity to improve search performance and exact symbolic recovery, our goal is to preserve multiple competitive expressions that differ in residual allocation.
Each expression is mapped to an archive cell defined by behavior descriptors such as the cluster attaining the largest mean absolute residual and structural characteristics like the number of nodes in the expression tree.
This enables DRSR to maintain multiple expressions with different residual patterns and structural characteristics, so that users can select an expression consistent with their domain knowledge using the behavior descriptors.
We evaluate DRSR on synthetic datasets with injected outliers, where a robust-loss-based quality metric improves robustness to outliers.
Furthermore, on a synthetic mixture dataset generated from two different underlying models, DRSR recovers accurate expressions for both relationships within a single run.
Finally, on a real-world astronomical dataset, we show that DRSR retrieves multiple candidate expressions consistent with models previously identified by domain experts.

% ============================================================ %
\section{Preliminaries} \label{sec:}
% ============================================================ %

% -------------------------------------------------- %
\subsection{Symbolic Regression} \label{ssec:SR}
% -------------------------------------------------- %
Symbolic regression (SR) aims to discover an explicit mathematical expression that describes the relationship between input variables and a target variable.
Given a dataset $\Data = \{ (\x_i,y_i) \}_{i=1}^n $, where $\x_i \in \mathbb{R}^d$ denotes an input vector and $y_i \in \mathbb{R}$ the corresponding output, the goal of SR is to identify a symbolic expression $f(\x)$ such that $f(\x_i)$ approximates $y_i$ well for all observations.

Unlike conventional regression methods that assume a fixed functional form, SR simultaneously searches both the structure and the parameters of the model.
Candidate expressions are typically constructed from a predefined function set (e.g., arithmetic operators and elementary functions) and a terminal set consisting of input variables and constants.
This flexibility allows SR to recover compact and interpretable models that can be directly inspected, simplified, and compared with existing domain knowledge.

A common approach to SR is genetic programming (GP)~\cite{gp}, where symbolic expressions are represented as tree structures and evolved through variation operators such as crossover and mutation.
The quality of an expression is evaluated using a loss function defined over the dataset, and search is guided toward expressions that achieve better predictive performance.
Mean squared error (MSE),
\begin{align}
    \mathcal{L}_{\textrm{MSE}}(f; \mathcal{D}) = \frac{1}{n} \sum_{i=1}^n \left( y_i - f(\x_i) \right)^2 \enspace,
\end{align}
and related metrics are widely used for this purpose, although alternative loss functions are also possible.

In practice, it is common for multiple symbolic expressions to achieve similar predictive accuracy on the same dataset while exhibiting different residual patterns across observations.
Such non-uniqueness arises from the combinatorial nature of the symbolic search space and the choice of loss function.
As a result, different expressions may explain different subsets of the data equally well, despite comparable aggregate error values.

% -------------------------------------------------- %
\subsection{Quality Diversity} \label{ssec:QD}
% -------------------------------------------------- %
Quality Diversity (QD)~\cite{cully2017quality} refers to a class of evolutionary search methods designed to discover a set of diverse, high-quality solutions within a single run. Unlike conventional optimization approaches that focus on identifying a single best solution with respect to a scalar objective, QD decomposes solution evaluation into \emph{quality}, which measures task performance, and \emph{behavior}, which is represented by user-defined descriptors capturing meaningful differences among solutions. By explicitly separating these two aspects, QD enables the preservation of multiple competitive solutions that differ in behavior while maintaining high performance.

A representative QD algorithm is MAP-Elites~\cite{MAP-Elites:2015}, which maintains an archive structured as a discretized grid over the behavior descriptor space.
Each cell in the archive stores the elite solution with the highest quality discovered for that region of the behavior space.
During the search, newly generated candidates are evaluated in terms of both quality and behavior and replace existing elites if they achieve higher quality within the same cell.
This archive-based mechanism facilitates the systematic exploration of the behavior space while retaining high-performing solutions.

QD methods have been applied to a range of domains where the availability of multiple alternative solutions is beneficial for analysis and interpretation.
By maintaining a diverse set of solutions with comparable quality but distinct behaviors, QD supports post hoc selection and analysis without requiring all user preferences to be encoded as optimization objectives.

% ============================================================ %
\section{Introduction of Robust Loss Functions into SR} \label{sec:robust}
% ============================================================ %
GP-based SR often evaluates candidate expressions by minimizing
the MSE.
While MSE is mathematically convenient and widely used, its squared penalty makes it highly sensitive to a small number of large residuals.
In this section, we introduce alternative loss functions derived from robust statistics that mitigate the influence of large residuals.
Additionally, we discuss the optimization challenges introduced by these losses, specifically the emergence of multiple solutions that differ in their outlier identification.

% -------------------------------------------------- %
\subsection{Robust Loss Functions} \label{ssec:robust-regression}
% -------------------------------------------------- %
To improve robustness, the loss function should penalize large residuals less severely than the quadratic growth of MSE. A standard alternative is the mean absolute error (MAE):
\begin{align}
    \Loss_\MAE (f; \Data) = \frac{1}{n} \sum_{i=1}^n |y_i - f(\x_i)|\enspace.
\end{align}
Since MAE grows linearly, a few extreme residuals cannot dominate the aggregate loss as strongly as under MSE.
For scenarios with many expected outliers, the median absolute error (MedAE) is often employed:
\begin{align}
    \Loss_\MedAE (f; \Data) = \mathrm{Median}\left( \Bigl\{ \left| y_i - f(\x_i) \right| ~ \middle| ~ i = 1, \ldots, n  \Bigr\} \right)\enspace.
\end{align}
Since MedAE depends on the median absolute residual, even arbitrarily large residuals in a minority of observations do not change the loss.
Equivalently, MedAE effectively bases the fit on the residuals of the best-fitting half of the observations, making it appropriate when severe contamination is expected.

While robust loss functions are standard tools in statistics, their use in GP-based SR is relatively limited.
For example, PySR~\cite{PySR:2023}, a widely used open-source SR tool, allows users to replace the default squared loss with an absolute-error loss (an $\ell_1$ loss, equivalent to MAE up to the constant factor $1/n$) as well as other robust losses such as the Huber loss~\cite{Robust-Huber:1992}.
However, in SR these options are often treated as configuration choices, and their implications for the search dynamics are not discussed in detail.

% -------------------------------------------------- %
\subsection{Landscape of Robust Loss Minimization} \label{ssec:landscape}
% -------------------------------------------------- %
\begin{figure}[t]
    \centering
    \includegraphics[width=0.9\linewidth]{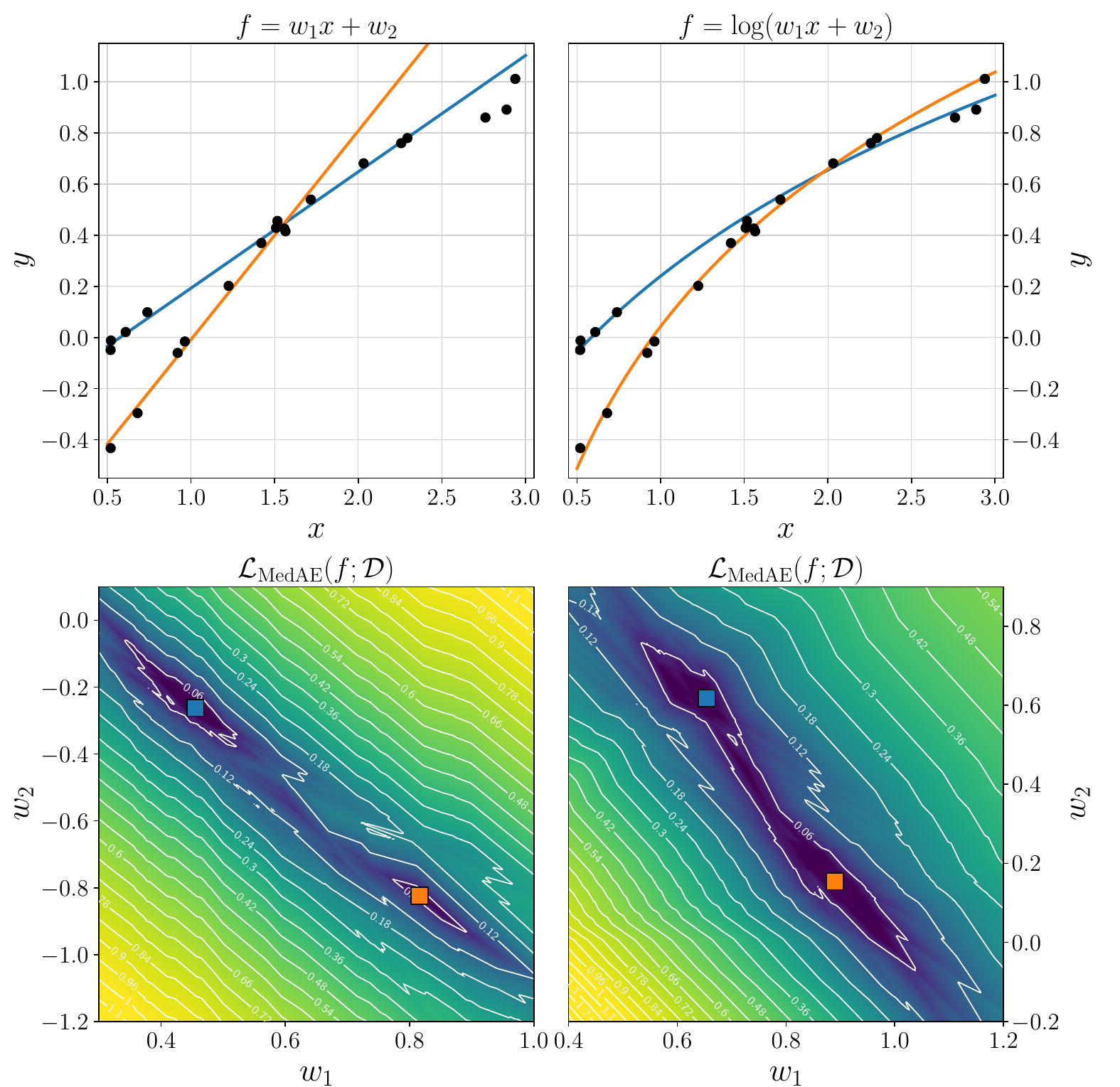}
    \vspace{-2mm}
    \caption{MedAE landscapes over the coefficients for two expression structures. The top panels show the example data points and the model outputs for two selected coefficient settings. The bottom panels show contour plots of $\Loss_{\MedAE}(f;\Data)$ on the $(w_1, w_2)$ plane. Blue and orange squares indicate the selected settings.}
    \label{fig:landscape}
\end{figure}

To make the issue concrete, Figure~\ref{fig:landscape} visualizes the MedAE objective as a function of the coefficients $(w_1, w_2)$ for two fixed expression structures.
For each case, the bottom panel shows $\Loss_\MedAE(f;\Data)$ over a grid of $(w_1,w_2)$ computed from the example data points, and the top panel plots the corresponding functions for two selected coefficient settings, using matching colors.
Even for the linear case $f=w_1x+w_2$, the MedAE landscape exhibits multiple basins of low loss, which is often observed and discussed in the field of robust regression~\cite{robust-local:2004, robust-local:2019}.
The same behavior is also observed for the nonlinear case $f=\log(w_1x+w_2)$, which is more representative of SR.
These results indicate that different basins correspond to expressions that fit different subsets of the observations well, i.e., they implicitly treat different observations as outliers.

This suggests two practical challenges when adopting robust losses in SR.
First, even when the expression structure is fixed, coefficient tuning can become a non-convex and multi-modal optimization problem.
As a result, local refinement methods that work well under squared error can be less reliable.
Second, during the search, several qualitatively different expressions can achieve similar robust loss values while exhibiting different residual patterns.
These challenges motivate two methodological choices in this work: the use of non-convex optimizers, such as covariance matrix adaptation evolution strategy (CMA-ES)~\cite{cmaes96, cmaes03}, for coefficient optimization and Quality-Diversity frameworks that preserve multiple promising expressions characterized by different residual patterns.

% ============================================================ %
\section{Proposed Framework} \label{sec:proposed}
% ============================================================ %
As discussed in the previous section, robust losses can yield multiple competitive solutions that differ mainly in which observations are fit well and where large residuals remain.
Because such residual allocation is often ambiguous and depends on domain knowledge, returning a single promising expression may hide alternative plausible explanations.
To address this issue, we propose \emph{diversified residual symbolic regression} (DRSR), a QD-based framework for SR that aims to produce a diverse set of competitive expressions characterized by residual-based behavior descriptors.

Algorithm~\ref{alg:DRSR} shows the overall DRSR procedure.
DRSR employs a grid archive $\Archive$ that stores candidate expressions according to the behavior descriptors defined below (lines~8 and 13).
DRSR then uniformly samples two expressions from $\Archive$ (line~3) and generates offspring by applying crossover and mutation operators commonly used in GP-based SR (lines~4 and~6).
To improve search efficiency, we further apply expression simplification (line~7) and CMA-ES–based coefficient optimization according to the user-specified loss $\Loss$ (lines~9--13), both described in detail below.

\subsection{Behavior Descriptors}

\paragraph{Outlier Cluster Index}
Let $\{\mathcal{C}_k\}_{k=1}^{K}$ be a partition of the dataset $\Data$ into $K$ clusters provided by the user.
In practice, a simple default is to rescale the input--output space and apply $k$-means~\cite{kmeans:1967}, but any domain-informed clustering is applicable.
We define the outlier cluster index as the cluster whose mean absolute residual is the largest:
\begin{align}
    \bb_{\mathrm{out}} (f; \Data) := \argmax_{k \in \{ 1,\ldots, K \}} \frac{1}{|\mathcal{C}_k|} \sum_{(\x, y) \in \mathcal{C}_k} | y - f(\x)| \enspace. \label{eq:bout}
\end{align}
Intuitively, $\bb_{\mathrm{out}}$ indicates which cluster the expression fits worst on average.
Because robust losses can downweight large residuals in different ways, two expressions with similar robust loss values may still assign large residuals to different regions of the data.
By indexing the archive with $\bb_{\mathrm{out}}$, DRSR explicitly preserves such alternative expressions.

\paragraph{Representation Power}
We use the number of nodes in the expression tree of $f$ as a structural descriptor, $\bb_{\mathrm{rep}} (f)$.
For example, the tree in Figure~\ref{fig:gp_tree} contains six nodes, hence $\bb_{\mathrm{rep}}(f)=6$.
GP-based SR with robust regression can yield either a compact expression that effectively treats certain observations as outliers, or a more expressive expression that also explains those observations.
By indexing the archive with $\bb_{\mathrm{rep}}(f)$, we explicitly preserve expressions across different capacity levels, enabling users to choose between simpler and more inclusive expressions.

\paragraph{Transcendental-function Count}
For a motivation similar to that of $\bb_{\mathrm{rep}}(f)$, we define $\bb_{\mathrm{trans}} (f)$ as the number of transcendental functions contained in $f$.
In Figure~\ref{fig:gp_tree}, the tree contains one $\exp$ node, thus $\bb_{\mathrm{trans}}(f)=1$.
This descriptor helps retain diverse expressions ranging from purely algebraic forms to those involving transcendental terms, enabling users to select expressions consistent with domain preferences.

\subsection{Simplification of Expressions}
For every generated offspring, we first perform constant folding to evaluate and replace sub-expressions composed solely of constants (line~7).
We then simplify the expressions using the \texttt{SymPy} library~\cite{SymPy:2017}.
This SymPy-based simplification includes grouping terms by $\x_i$ (\texttt{collect}), canceling common factors between numerators and denominators (\texttt{cancel}), and combining exponents (\texttt{powsimp}).
We apply these transformations provided that they do not increase the number of nodes.

\subsection{Coefficient Optimization}
\begin{figure}[t]
    \centering
    \includegraphics[width=0.5\linewidth]{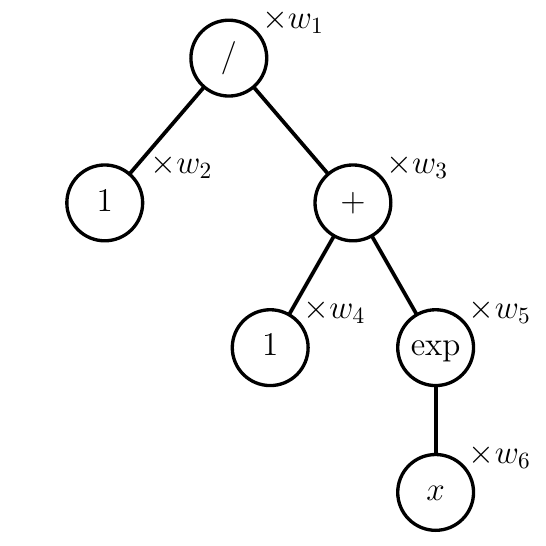}
    \vspace{-1mm}
    \caption{Parameterized GP tree for $\frac{1}{1+\exp(x)}$.}
    \label{fig:gp_tree}
\end{figure}

We parameterize GP trees by introducing learnable multiplicative weights that scale node outputs, following the general idea of parametrizing GP trees in~\cite{parameterized:2023}.
Unlike~\cite{parameterized:2023}, which places parameters on edge connections, we use node-wise weights and also parameterize the root output to support single-node expressions.
For example, Figure~\ref{fig:gp_tree} shows the parameterized GP tree based on the structure of $\frac{1}{1+\exp(x)}$.
With weights $\{w_i\}$, the instantiated expression becomes
\begin{align}
    w_1 \cdot \frac{w_2 \cdot 1}{w_3 \cdot \Bigl( w_4 \cdot 1 + w_5 \cdot \exp( w_6 \cdot x ) \Bigr)} \enspace.
\end{align}
As discussed in Section~\ref{ssec:landscape}, when using a robust loss function such as MedAE, the coefficient landscape can become non-convex and multimodal, making gradient-based tuning unreliable or inapplicable.
Therefore, we optimize coefficients with CMA-ES (lines~11 and 12) and store the generated candidate expressions $\hat{f}_1, \ldots, \hat{f}_\lambda$ in $\Archive$.
As a result, $\Archive$ can later return multiple coefficient-tuned expressions corresponding to different locally refined coefficient configurations.

\begin{algorithm}[t]
    \caption{DRSR Framework}
    \begin{algorithmic}[1]
        \STATE \textbf{input:} $\Archive$ (archive)
        \WHILE{termination conditions are not met}
        \STATE $f_1, f_2 \leftarrow \Archive.\text{selection}()$
        \STATE $f_1, f_2 \leftarrow \text{crossover}(f_1, f_2)$
        \FOR{$f \in \{f_1, f_2\}$}
        \STATE $f \leftarrow \text{mutation}(f)$
        \STATE $f \leftarrow \text{simplification}(f)$
        \STATE $\Archive.\text{update}(f)$
        \STATE $\textrm{ES} \leftarrow \text{init\_ES}(f)$ \COMMENT{based on coefficients of $f$}
        \WHILE{ES conditions are not met}
        \STATE $\hat{f}_1, \ldots, \hat{f}_\lambda \leftarrow \text{ES}.\text{ask}()$ \COMMENT{$f$ with coefficients set by ES}
        \STATE $\text{ES}.\text{tell}( \mathcal{L}(\hat{f}_1 ; \Data), \ldots,  \mathcal{L}(\hat{f}_\lambda ; \Data) )$ \COMMENT{minimize the loss}
        \STATE $\Archive.\text{update}( \hat{f}_1, \ldots, \hat{f}_\lambda )$
        \ENDWHILE
        \ENDFOR
        \ENDWHILE
    \end{algorithmic}
    \label{alg:DRSR}
\end{algorithm}

% ============================================================ %
\section{Evaluation on Synthetic Datasets} \label{sec:exp1}
% ============================================================ %
In this section, we evaluate the effectiveness of the proposed framework, DRSR, on synthetic datasets.

\subsection{General Setting}
\paragraph{Representation and GP Approach}
DRSR follows a GP approach, iteratively generating and evaluating candidate expressions across generations, where each candidate is represented as an expression tree composed of a function set and a terminal set.
The function set is $\{+,-,\times,/,\log,\exp\}$, and the terminal set consists of the input variable $\x$ and the constant 1.
To constrain expression complexity, we set the maximum number of nodes to 20 and the maximum tree depth to 17.
Furthermore, a coefficient initialized to 1 is assigned to each node output in the tree structure.
When evaluating an expression by substituting the input $\x$, we set the residual for that observation to $10^6$ if (i) the argument of $\log$ is negative, (ii) the absolute value of the denominator in $/$ is smaller than $10^{-12}$, or (iii) the absolute value of the argument of $\exp$ exceeds $100$.
Rather than using protected functions, this design enables the use of functions with mathematically exact definitions.
The initial population consists of $1000$ individuals generated using ramped half-and-half~\cite{koza1992}.
Offspring are generated by selecting two parents and applying subtree crossover with probability $0.9$, followed by subtree mutation with probability $0.1$.

\paragraph{Behavior Descriptors and Archive}
In our experiments, we consider three behavior descriptors: (i) the \emph{outlier cluster index} defined in \cref{eq:bout}, (ii) the \emph{representation power} measured by the number of nodes in the expression tree, and (iii) the \emph{transcendental-function count}, i.e., the number of $\log$ and $\exp$ nodes in the expression tree.
Before the search, we apply an affine rescaling to each dimension of the dataset so that all values lie in $[0,1]$ only for clustering, and then run $k$-means with $K=10$ to obtain clusters $\{\mathcal{C}_k\}_{k=1}^{K}$.
We use a unit-resolution grid so that each integer value corresponds to a single archive cell, and restrict archive ranges of the representation power and the transcendental-function count to $[1,20]$ and $[0,4]$, respectively. If the transcendental-function count exceeds the upper bound, we map it to the boundary value $4$.

\paragraph{Coefficient Optimization}
For the coefficient optimization by CMA-ES, we set the population size to 10 and the number of generations to 20. The initial step size is set to 1, and the initial mean vector is defined by the initial coefficient values.

% -------------------------------------------------- %
\subsection{Effectiveness of Robust Loss Functions against Outliers} \label{ssec:exp1:robust_loss}
% -------------------------------------------------- %
In this subsection, we demonstrate that employing robust loss functions described in \Cref{ssec:robust-regression} allows DRSR to effectively mitigate the impact of outliers.

\paragraph{Experimental Settings}
We employ four Nguyen benchmark functions~\cite{Uy:2011} listed in \Cref{tab:nguyen_benchmarks}.
Nguyen-1 and Nguyen-7 are univariate functions, where $\x=x$, while Nguyen-11 and Nguyen-12 are bivariate functions, where $\x=(x_1, x_2)$.
For each benchmark, we generate a dataset $\Data$ composed of clean base data points and noisy outliers: $\Data = \Data_{\mathrm{base}} \cup \Data_{\mathrm{noise}}$.
The subsets are defined as $\Data_{\mathrm{base}} = \{(\x_i, y_i)\}_{i=1}^{n_\mathrm{base}}$ and $\Data_{\mathrm{noise}} = \{(\x_i, y_i + \epsilon_i)\}_{i=1}^{n_\mathrm{noise}}$, where inputs $\x_i$ are sampled uniformly from the specified domain and $\epsilon_i \sim \mathcal{N}(0, 1)$ represents the additive noise.
In this experiment, we fix the number of base points at $n_{\mathrm{base}} = 20$ and vary the number of outliers as $n_{\mathrm{noise}} \in \{5, 20\}$.
\begin{table}[tb]
    \centering
    \caption{
        Nguyen benchmarks used in the experiments.
    }
    \label{tab:nguyen_benchmarks}
    \begin{tabular}{l l l}
        \toprule
        Name      & Expression                                  & Domain     \\
        \midrule
        Nguyen-1  & $y= x^3 + x^2 + x$                          & $[-1, 1]$  \\
        Nguyen-7  & $y= \log(x+1) + \log(x^2+1)$                & $[0, 2]$   \\
        Nguyen-11 & $y= x_1^{x_2}$                              & $[0, 1]^2$ \\
        Nguyen-12 & $y= x_1^4 - x_1^3 + \frac{1}{2}x_2^2 - x_2$ & $[0, 1]^2$ \\
        \bottomrule
    \end{tabular}
\end{table}

We apply DRSR using three loss functions: $\Loss_\mathrm{MedAE}$, $\Loss_\mathrm{MAE}$, and $\Loss_\mathrm{MSE}$, where $\Loss_\mathrm{MedAE}$ and $\Loss_\mathrm{MAE}$ are robust loss functions.
The fitness is defined as $1/(1 + \Loss(f;\Data))$, formulating the task as a maximization problem.
For each loss function, we conduct 10 independent trials of DRSR with a fitness evaluation budget of $2 \times 10^6$.

\paragraph{Metrics}
To assess how well DRSR with each loss function captures the underlying relationship, we evaluate the accuracy of the best expression in the current generation on the base subset of the dataset as $1/(1+\Loss_{\mathrm{MSE}}(f;\Data_{\mathrm{base}}))$.

\begin{figure*}[t]
    \centering
    \includegraphics[width=0.93\linewidth]{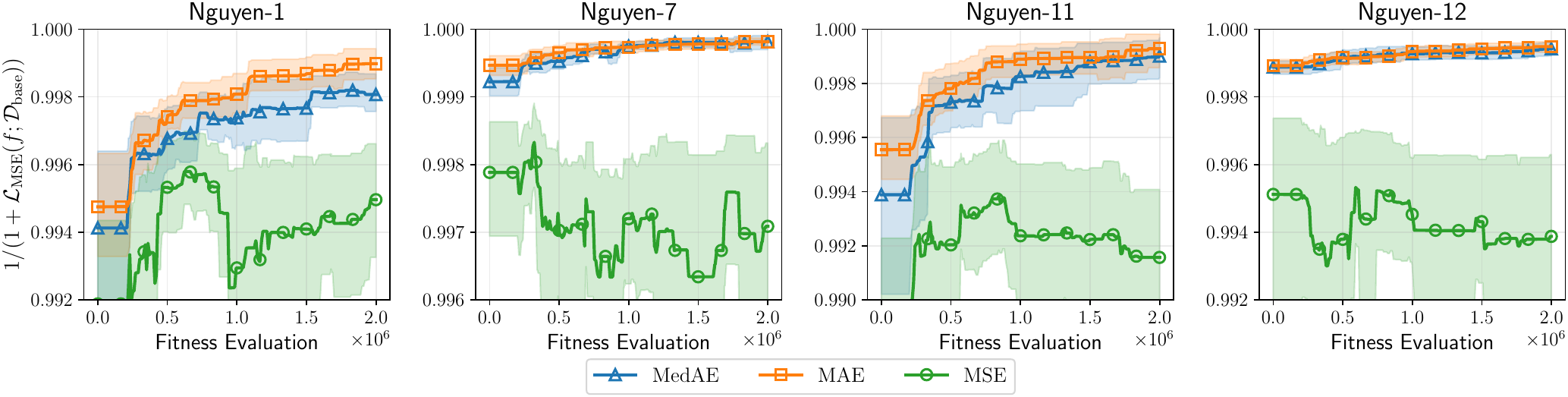}
    \caption{
        Comparison of trajectories of the best predictive accuracy in each generation of DRSR on Nguyen benchmarks with $n_{\mathrm{noise}} = 5$ under three loss functions: $\Loss_\mathrm{MedAE}$, $\Loss_\mathrm{MAE}$, and $\Loss_\mathrm{MSE}$.
        The solid curves show the mean accuracy, and the shaded regions show the 95\% bootstrap confidence intervals over 10 independent trials.
        The accuracy is defined as $1/(1 + \Loss_\mathrm{MSE}(f;\Data_\mathrm{base}))$.
    }
    \label{fig:nguyen_5noise}
\end{figure*}

\begin{figure*}[t]
    \centering
    \includegraphics[width=0.93\linewidth]{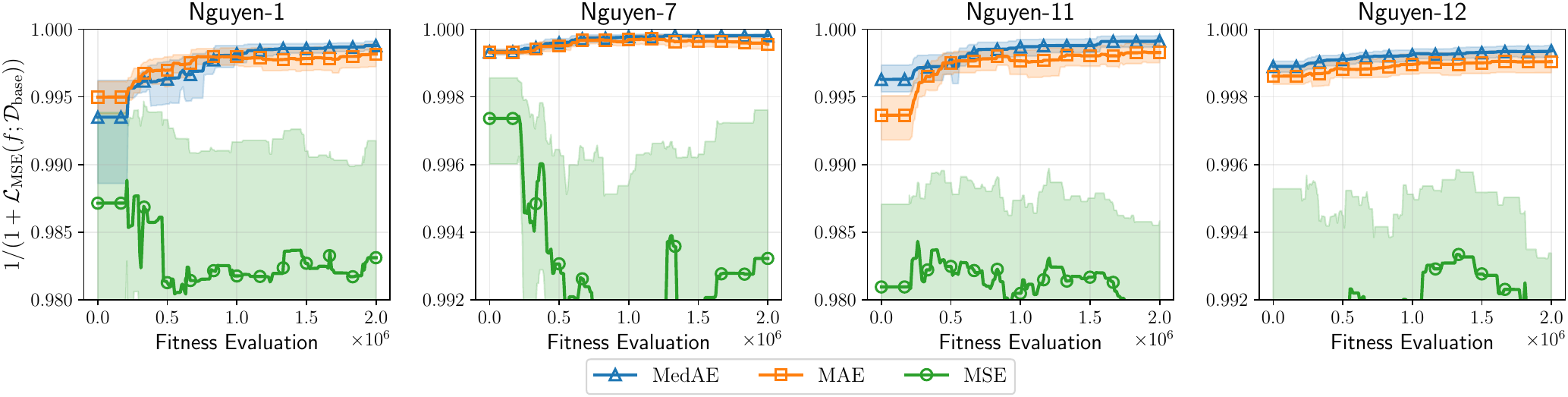}
    \caption{
        Comparison of trajectories of the best predictive accuracy in each generation of DRSR on Nguyen benchmarks with $n_{\mathrm{noise}} = 20$ under three loss functions: $\Loss_\mathrm{MedAE}$, $\Loss_\mathrm{MAE}$, and $\Loss_\mathrm{MSE}$.
        The solid curves show the mean accuracy, and the shaded regions show the 95\% bootstrap confidence intervals over 10 independent trials.
        The accuracy is defined as $1/(1 + \Loss_\mathrm{MSE}(f;\Data_\mathrm{base}))$.
    }
    \label{fig:nguyen_20noise}
\end{figure*}

\paragraph{Results}
\Cref{fig:nguyen_5noise} and \Cref{fig:nguyen_20noise} show the trajectories of the accuracy of the best expression in the current generation on each benchmark using different loss functions for $n_{\mathrm{noise}} = 5$ and $20$, respectively.
In these plots, the solid curves represent the mean accuracy, while the colored shaded regions indicate the 95\% bootstrap confidence intervals.

As shown in \Cref{fig:nguyen_5noise}, when $n_{\mathrm{noise}} = 5$, DRSR employing robust loss functions (MedAE and MAE) achieved higher accuracy than DRSR using MSE.
Notably, in this setting, MAE demonstrated accuracy comparable to or superior to MedAE.
Conversely, \Cref{fig:nguyen_20noise} shows that when the number of outliers is increased to $n_{\mathrm{noise}} = 20$, while both robust loss functions still outperform MSE, MedAE achieves accuracy comparable to or superior to MAE.
These results demonstrate the effectiveness of robust loss functions in handling noisy datasets.
Furthermore, the findings suggest that as the number of outliers increases, MedAE tends to be more effective than MAE.
Similar trends were also observed for SR and MOSR, and the corresponding results are provided in Appendix A.

% -------------------------------------------------- %
\subsection{Discovering Diverse Expressions in a Mixture Dataset} \label{ssec:}
% -------------------------------------------------- %
In this subsection, we demonstrate that, on a synthetic mixture dataset, DRSR discovers a more diverse set of high-quality expressions than baseline methods while capturing the two underlying relationships.

\paragraph{Experimental Settings}
We employ a synthetic mixture dataset $\Data_{\mathrm{mixed}}$, motivated by price--demand relationships for a single product under heterogeneous consumer behaviors~\cite{Bodea:2014, Phillips:2021}.
We sample 40 univariate inputs $x_i$ uniformly from the interval $[0, 10]$.
The dataset is constructed as $\Data_{\mathrm{mixed}} = \Data_{\mathrm{linear}} \cup \Data_{\mathrm{logistic}}$, where the target variable $y$ is generated by a probabilistic mixture model defined as follows:
\begin{align}
    y = & \left\{
    \begin{alignedat}{2}
        &a - bx \ && \quad \text{for } \Data_{\mathrm{linear}}, \ \text{w.p. } 0.5 \\
        &\frac{1}{1+\exp(c+dx)} \ && \quad \text{for } \Data_{\mathrm{logistic}}, \ \text{w.p. } 0.5
    \end{alignedat}
    \right. \enspace.
\end{align}
Here, the parameters are set to $a=1, b=0.1, c=-4$, and $d=1.6$.
The linear component $\Data_{\mathrm{linear}}$ represents a demand that decreases linearly with price $x$.
In contrast, the logistic component $\Data_{\mathrm{logistic}}$ represents a nonlinear saturated demand with high sensitivity in the intermediate price range.
This mixture model simulates a scenario where different purchasing behaviors coexist within the same dataset.

We evaluate the performance of DRSR in comparison with two baseline approaches: standard single-objective symbolic regression (SR) and multiobjective symbolic regression (MOSR), where MOSR simultaneously maximizes the fitness of expressions and minimizes the number of nodes in the expression tree.
To ensure a fair comparison, both baselines utilize the same GP-based tree representation, offspring generation operators, coefficient optimization, and simplification procedures as employed in DRSR.
To robustly identify the underlying relationships in the mixture dataset, all methods employ the MedAE-based fitness $F(f;\Data_\mathrm{mixed})=1/(1 + \Loss_\mathrm{MedAE}(f;\Data_\mathrm{mixed}))$.
We conduct 10 independent trials of each method with a fitness evaluation budget of $2 \times 10^6$.

The specific configurations for the baselines are as follows:
For SR, the population size is set to 1000.
Parents are chosen via tournament selection with a tournament size of 3, and the next generation consists of the offspring excluding the worst solution among them, supplemented by the best solution from the current generation.
For MOSR, the population size is also set to 1000.
Parents are chosen via tournament selection, and the next generation is determined via truncation selection, where both selections are based on the non-dominated sorting of NSGA-II~\cite{NSGA-II:2002}.

\paragraph{Metrics}
We track several metrics to evaluate overall and \linebreak component-wise predictive performance, as well as solution diversity.
We report the \emph{best fitness} achieved during the search.
To measure how well each method captures the two underlying relationships,
we compute component-wise accuracies for the linear and logistic subsets, defined as $1/(1+\Loss_\mathrm{MSE}(f;\Data_\mathrm{linear}))$ and $1/(1+\Loss_\mathrm{MSE}(f;\Data_\mathrm{logistic}))$, respectively.

To quantify diversity, we report \emph{coverage}, defined as the fraction of occupied cells in the archive, and the \emph{QD-Score}, defined as the sum of fitness values of all elite solutions divided by the total number of cells in the archive.
For SR and MOSR, which do not maintain an archive, these metrics are computed using a temporary archive constructed from the current population.

Finally, we report the \emph{hypervolume (HV)} to assess the trade-off between accuracy and simplicity.
HV is computed over the maintained solution set (the archive for DRSR and the current population for SR and MOSR), considering two objectives: maximizing fitness and minimizing the number of nodes, with a reference point set to $(0, 20)$.

\begin{figure*}[t]
    \centering
    \includegraphics[width=0.6\linewidth]{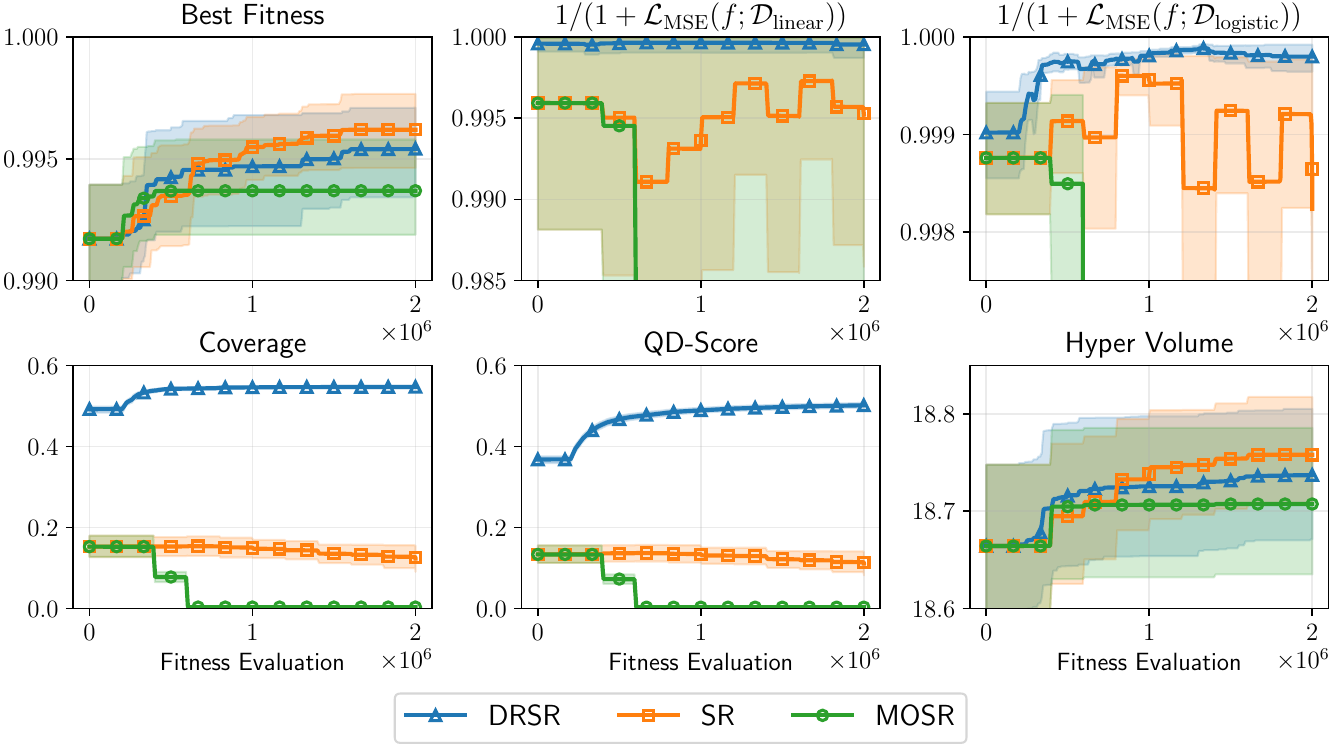}
    \vspace{-2mm}
    \caption{
        Evaluation metric trajectories of DRSR, SR, and MOSR on the synthetic mixture dataset.
        We track best fitness, accuracies on the linear and logistic components, coverage, QD-Score, and HV.
        The solid curves show the mean values of the corresponding metrics, and shaded regions show the 95\% bootstrap confidence intervals over 10 independent trials.
    }
    \label{fig:mixture:performance}
\end{figure*}

\paragraph{Results}
\Cref{fig:mixture:performance} shows the trajectories of the evaluation metrics for DRSR, SR, and MOSR on the mixture dataset.
In these plots, the solid curves represent the mean values of the corresponding metrics, and the colored shaded regions indicate the 95\% bootstrap confidence intervals.

As shown in \Cref{fig:mixture:performance}, in terms of best fitness and HV, DRSR is comparable to or inferior to SR, while remaining comparable to or superior to MOSR.
In contrast, for the accuracies on the linear and logistic components, DRSR is comparable to or superior to both SR and MOSR.
Furthermore, in terms of coverage and QD-Score, DRSR significantly outperforms both SR and MOSR.
A possible reason for the early convergence of MOSR to low coverage and QD-Score is that the dataset admits simple yet highly accurate expressions, such as linear ones, which may dominate the population under the accuracy--simplicity trade-off.

\begin{figure*}[t]
    \centering
    \includegraphics[width=0.8\linewidth]{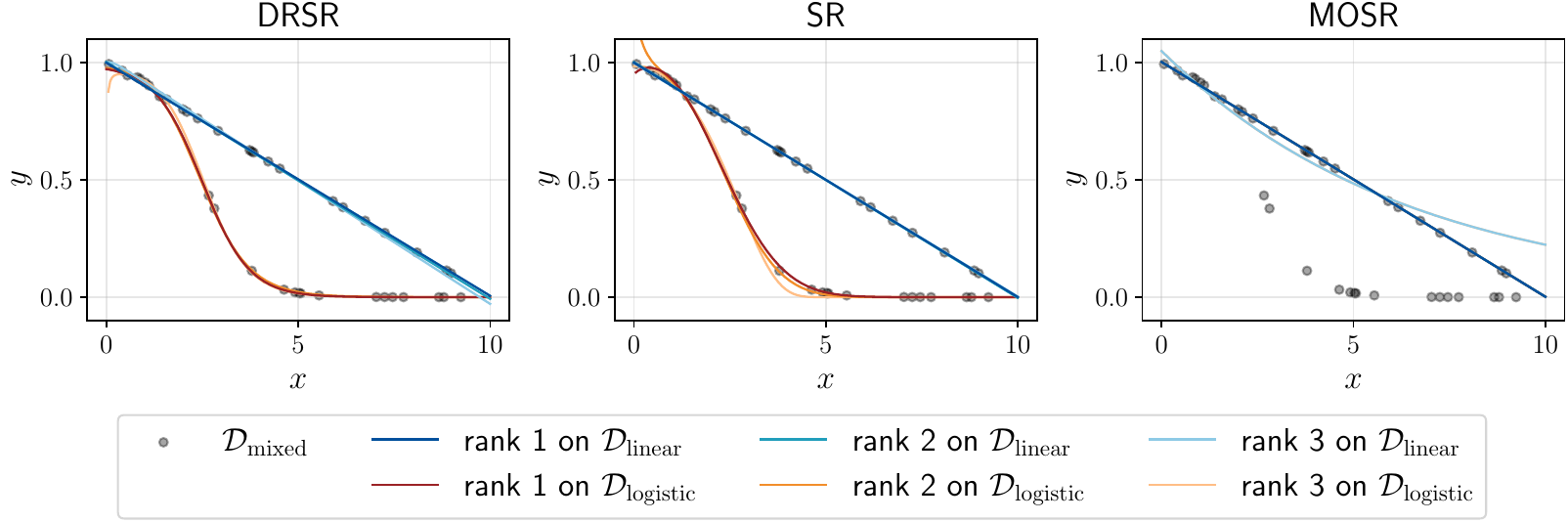}
    \vspace{-1.5mm}
    \caption{
        Predictive behaviors of the top 3 expressions from DRSR, SR, and MOSR with the highest component-wise accuracies for the linear dataset $\Data_\mathrm{linear}$ and logistic dataset $\Data_\mathrm{logistic}$, respectively.
        The expressions are selected from the final solution set of a trial, and the scatter points represent the mixture dataset $\Data_\mathrm{mixed}$.
        For MOSR, the curves of the top 3 expressions on $\Data_\mathrm{logistic}$ are not separately visible because they are identical to those on $\Data_\mathrm{linear}$.
    }
    \label{fig:mixture:xy}
\end{figure*}

\Cref{fig:mixture:xy} shows the $x$--$y$ plots of the top 3 expressions selected from the final solution set of a trial according to the component-wise accuracies for the linear and logistic components, defined as $1/(1+\Loss_\mathrm{MSE}(f;\Data_\mathrm{linear}))$ and $1/(1+\Loss_\mathrm{MSE}(f;\Data_\mathrm{logistic}))$, respectively.
Notably, in this trial, DRSR identified an expression with the exact symbolic structure of the logistic function as the best solution for the logistic component.
In contrast, neither SR nor MOSR produced an expression with the exact symbolic structure of the logistic function among their top three candidates in any trial.
For the linear component, all three methods identified an expression with the exact symbolic structure of the linear function as the best solution in this trial.
The explicit forms of all DRSR expressions shown in \Cref{fig:mixture:xy} are provided in Appendix B.1.

Collectively, these results suggest that DRSR is more effective than SR and MOSR in discovering a diverse set of expressions that accurately capture multiple underlying relationships while exhibiting diverse residual patterns.

% ============================================================ %
\section{Evaluation on a Real-World Astronomical Dataset} \label{sec:exp2}
% ============================================================ %
In this section, we evaluate DRSR on a real-world astronomical dataset and examine whether it can discover multiple expressions consistent with known physical relationships.

\paragraph{Experimental Settings}
We use the DEBCat stellar dataset~\cite{DEBCat:2014}, a catalog of the physical properties of well-studied detached eclipsing binaries\footnote{Accessed via VizieR at \url{https://vizier.cds.unistra.fr/viz-bin/VizieR?-source=V/152}}.
Since the catalog is not restricted to main-sequence stars, it may include systems at different stages of stellar evolution.
In this experiment, the input variable $x$ is the stellar mass (in solar units $\mathrm{M}_\odot$), and the target variable $y$ is the luminosity (in solar units $\mathrm{L}_\odot$).
The original catalog records both variables as base-10 logarithms, which we convert to linear units before the search.
We then normalize both $x$ and $y$ to $[0, 1]$ by an affine transformation.
We conduct 10 independent trials of DRSR on this dataset using the MedAE-based fitness $F(f;\Data)=1/(1+\Loss_{\mathrm{MedAE}}(f;\Data))$.
The remaining experimental settings, including the GP representation and hyperparameters, are the same as those in \Cref{sec:exp1}.

In astrophysics, the mass--luminosity relation of main-sequence stars is commonly modeled by a power law, $y \propto x^{\alpha}$~\cite{Eker:2018}.
Within our symbolic search space, this relation is represented by the structure $a \exp(b \log(c x))$, which is mathematically equivalent to $A x^b$.
The challenge, therefore, is whether DRSR can identify expressions with this structure from a catalog that is not restricted to main-sequence stars.

\begin{figure}[t]
    \centering
    \includegraphics[width=0.8\linewidth]{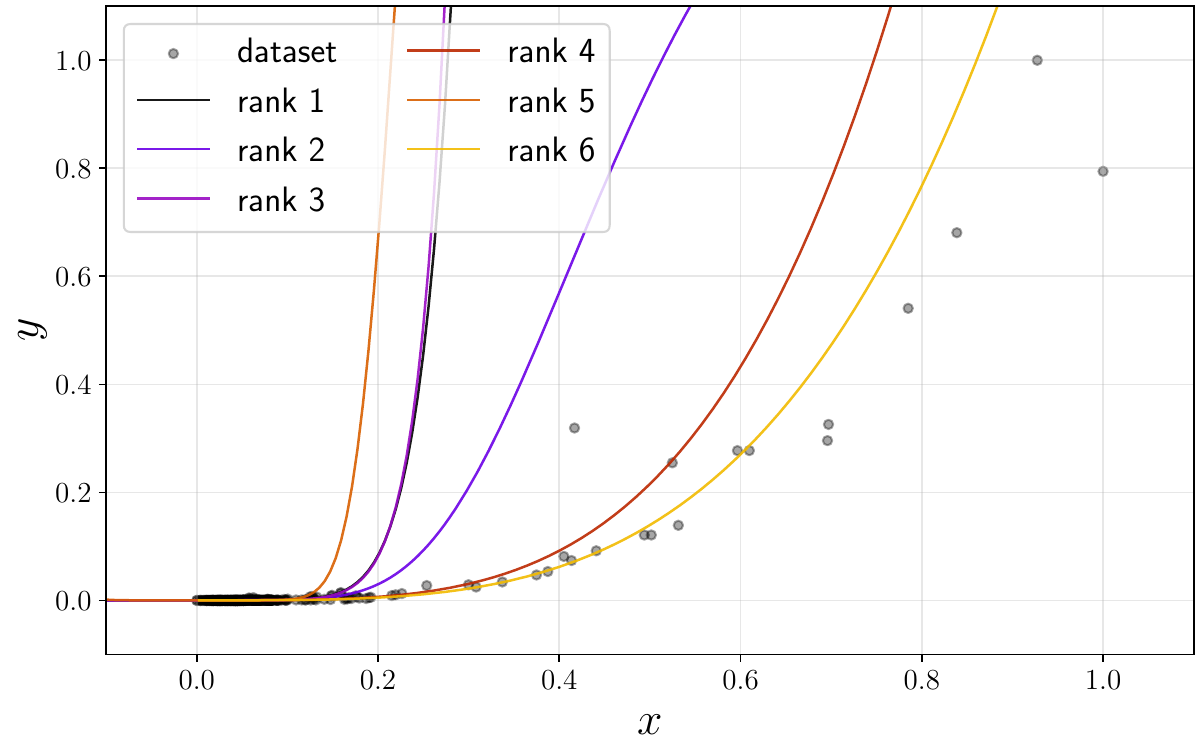}
    \vspace{-1.5mm}
    \caption{
        Top 6 expressions found by DRSR for the stellar mass--luminosity dataset.
        The expressions are selected from an archive region defined by $\bb_{\mathrm{rep}}(f)\in [1, 5]$ and $\bb_{\mathrm{trans}}(f)\in [0, 2]$ to simulate expert-guided filtering.
        The 4th- and 6th-ranked expressions have the power-law form $y = A x^b$.
    }
    \label{fig:astronomy:xy}
\end{figure}

\begin{figure}[t]
    \centering
    \includegraphics[width=0.8\linewidth]{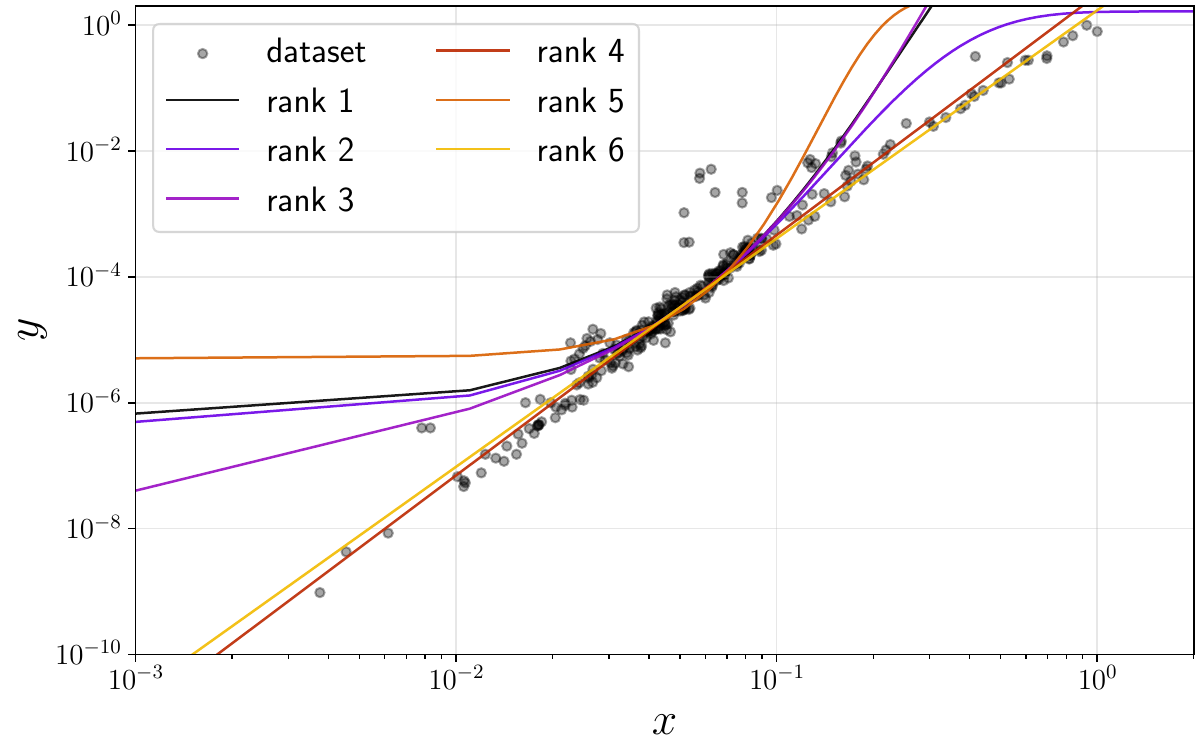}
    \vspace{-1.5mm}
    \caption{
        Log--log plots of the same expressions shown in \Cref{fig:astronomy:xy}.
        The 4th- and 6th-ranked expressions, which have the power-law form $y = A x^b$, appear as straight lines.
    }
    \label{fig:astronomy:logxlogy}
\end{figure}

\paragraph{Results}
In a practical application scenario, we anticipate that domain experts would leverage their prior knowledge to focus on specific regions of the archive rather than examining it in its entirety.
To simulate such expert-guided filtering, we restrict our analysis to expressions with representation power in $[1, 5]$ and transcendental-function count in $[0, 2]$.
\Cref{fig:astronomy:xy} shows the $x$--$y$ plots of the top 6 expressions with the highest fitness in this subset for one trial; their explicit forms are provided in Appendix B.2.
Notably, in this trial, the expressions ranked 4th and 6th have the structure $a\exp(b\log(c x))$, consistent with a mass--luminosity relation.
We also visualize the predictive behaviors of these expressions on logarithmic axes in \Cref{fig:astronomy:logxlogy} to confirm that they follow a power-law relation.
As shown in \Cref{fig:astronomy:logxlogy}, the trajectories of the 4th- and 6th-ranked expressions appear as straight lines.

\begin{figure}[t]
    \centering
    \includegraphics[width=0.8\linewidth]{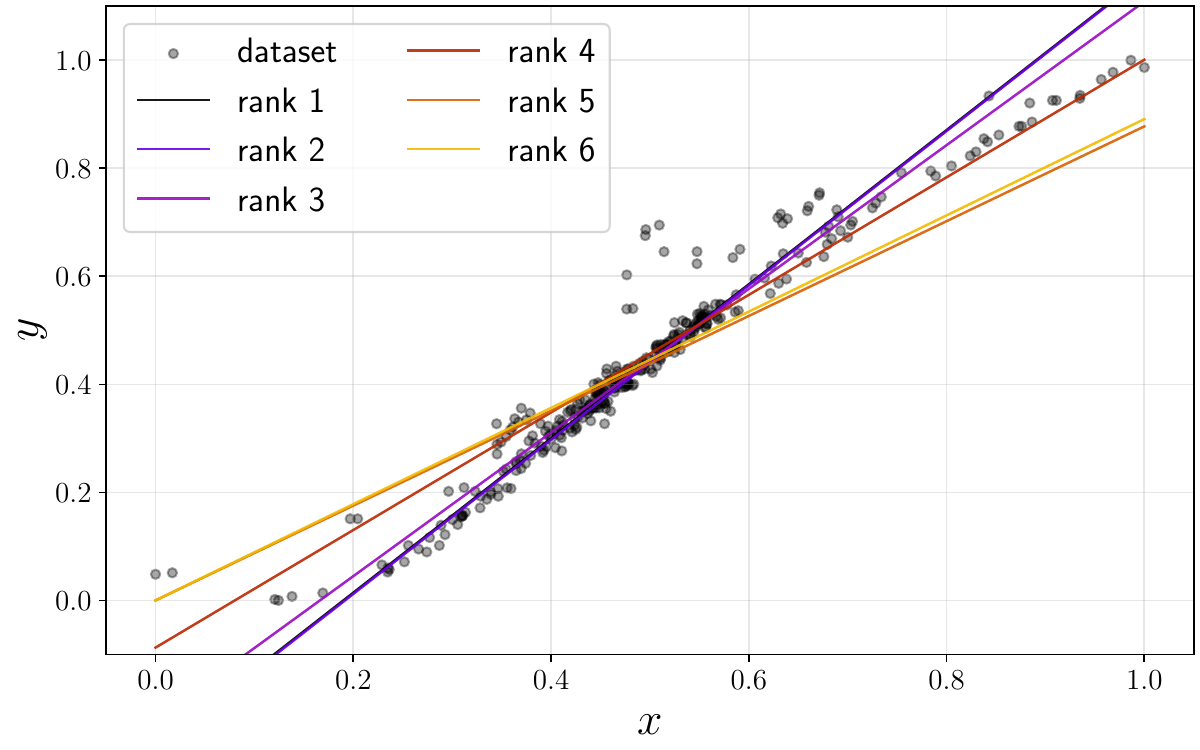}
    \vspace{-1.5mm}
    \caption{
        Top 6 expressions found by DRSR using the log-transformed dataset.
        The expressions are selected from an archive region defined by $\bb_{\mathrm{rep}}(f)\in [1,3]$ and $\bb_{\mathrm{trans}}(f)=0$.
    }
    \label{fig:astronomy:search_logxlogy}
\end{figure}

In addition to the direct search in the non-log-transformed space, we conduct a supplementary experiment on a log-transformed dataset to examine whether search in the transformed space yields simpler and more diverse expressions for the underlying relations.
In this setting, we first apply a logarithmic transformation to both $x$ and $y$ and then normalize them to $[0, 1]$ by an affine transformation.
We restrict our analysis to archive cells with representation power in $[1,3]$ and no transcendental functions.
\Cref{fig:astronomy:search_logxlogy} shows the $x$--$y$ plots of the top 6 expressions with the highest fitness in this subset from one trial.
The explicit forms of these expressions are provided in Appendix B.2, where we show that some of them are consistent with the piecewise classical mass--luminosity relations reported by \citet{Eker:2018}.

% ============================================================ %
\section{Conclusion} \label{sec:conclusion}
% ============================================================ %
In this paper, we proposed diversified residual symbolic regression (DRSR) to address the limitations of standard SR when facing outliers and data heterogeneity. By integrating robust statistical loss functions with the Quality-Diversity paradigm, DRSR utilizes residual-based behavior descriptors to maintain a diverse archive of high-quality expressions. This approach allows the search to preserve alternative hypotheses that differ in their residual patterns, rather than prematurely converging to a single compromise solution.

We evaluate DRSR on synthetic datasets with injected outliers, where a robust-loss-based quality metric improves robustness to outliers.
Furthermore, on synthetic mixture datasets, DRSR recovers multiple distinct underlying structures within a single run, and on a real-world astronomical dataset, it retrieves multiple candidate expressions consistent with known mass--luminosity relations.
We believe that DRSR provides a set of candidate expressions that can help domain experts identify those most consistent with their domain knowledge.

\begin{acks}
    The authors used generative AI tools for assistance with English writing and experimental code implementation.
\end{acks}

\bibliographystyle{ACM-Reference-Format}
\bibliography{reference}

@article{NSGA-II:2002,
author={Deb, K. and Pratap, A. and Agarwal, S. and Meyarivan, T.},
journal={IEEE Transactions on Evolutionary Computation},
title={A fast and elitist multiobjective genetic algorithm: NSGA-II},
year={2002},
volume={6},
number={2},
pages={182-197},
doi={10.1109/4235.996017}
}

@article{Eker:2018,
   title={Interrelated main-sequence mass–luminosity, mass–radius, and mass–effective temperature relations},
   volume={479},
   ISSN={1365-2966},
   url={http://dx.doi.org/10.1093/mnras/sty1834},
   DOI={10.1093/mnras/sty1834},
   number={4},
   journal={Monthly Notices of the Royal Astronomical Society},
   publisher={Oxford University Press (OUP)},
   author={Eker, Z and Bakış, V and Bilir, S and Soydugan, F and Steer, I and Soydugan, E and Bakış, H and Aliçavuş, F and Aslan, G and Alpsoy, M},
   year={2018},
   month=jul, pages={5491–5511} }

@misc{DEBCat:2014,
      title={The DEBCat detached eclipsing binary catalogue}, 
      author={John Southworth},
      year={2014},
      eprint={1411.1219},
      archivePrefix={arXiv},
      primaryClass={astro-ph.SR},
      url={https://arxiv.org/abs/1411.1219}, 
}

@article{Uy:2011,
author = {Uy, Nguyen Quang and Hoai, Nguyen Xuan and O'Neill, Michael and McKay, R. I. and Galv{\'a}n-L{\'o}pez, Edgar},
title = {Semantically-based crossover in genetic programming: application to real-valued symbolic regression},
journal = {Genetic Programming and Evolvable Machines},
year = {2011},
volume = {12},
number = {2},
pages = {91--119},
doi = {10.1007/s10710-010-9121-2},
}

@article{SymPy:2017,
     title = {SymPy: symbolic computing in Python},
     author = {Meurer, Aaron and Smith, Christopher P. and Paprocki, Mateusz and \v{C}ert\'{i}k, Ond\v{r}ej and Kirpichev, Sergey B. and Rocklin, Matthew and Kumar, AMiT and Ivanov, Sergiu and Moore, Jason K. and Singh, Sartaj and Rathnayake, Thilina and Vig, Sean and Granger, Brian E. and Muller, Richard P. and Bonazzi, Francesco and Gupta, Harsh and Vats, Shivam and Johansson, Fredrik and Pedregosa, Fabian and Curry, Matthew J. and Terrel, Andy R. and Rou\v{c}ka, \v{S}t\v{e}p\'{a}n and Saboo, Ashutosh and Fernando, Isuru and Kulal, Sumith and Cimrman, Robert and Scopatz, Anthony},
     year = 2017,
     month = jan,
     volume = 3,
     pages = {e103},
     journal = {PeerJ Computer Science},
     issn = {2376-5992},
     url = {https://doi.org/10.7717/peerj-cs.103},
     doi = {10.7717/peerj-cs.103}
    }

@book{gp,
  author    = {Koza, John R.},
  title     = {Genetic programming: on the programming of computers by means of natural selection},
  year      = {1992},
  isbn      = {0262111705},
  publisher = {MIT Press},
  address   = {Cambridge, MA, USA}
}

@inproceedings{koza1992,
  title={Evolution of subsumption using genetic programming},
  author={Koza, John R},
  booktitle={Proceedings of the first European conference on artificial life},
  pages={110--119},
  year={1992},
  organization={MIT Press Cambridge, MA, USA}
}

@inproceedings{ransacgp,
  author    = {L{\'o}pez, Uriel
               and Trujillo, Leonardo
               and Martinez, Yuliana
               and Legrand, Pierrick
               and Naredo, Enrique
               and Silva, Sara},
  editor    = {McDermott, James
               and Castelli, Mauro
               and Sekanina, Lukas
               and Haasdijk, Evert
               and Garc{\'i}a-S{\'a}nchez, Pablo},
  title     = {RANSAC-GP: Dealing with Outliers in Symbolic Regression with Genetic Programming},
  booktitle = {Genetic Programming},
  year      = {2017},
  publisher = {Springer International Publishing},
  pages     = {114--130},
  isbn      = {978-3-319-55696-3}
}

@article{SR-physics:2009,
author = {Michael Schmidt  and Hod Lipson },
title = {Distilling Free-Form Natural Laws from Experimental Data},
journal = {Science},
volume = {324},
number = {5923},
pages = {81-85},
year = {2009},
doi = {10.1126/science.1165893},
URL = {https://www.science.org/doi/abs/10.1126/science.1165893},
eprint = {https://www.science.org/doi/pdf/10.1126/science.1165893}
}

@article{SR-materials:2019,
title={Symbolic regression in materials science}, volume={9}, DOI={10.1557/mrc.2019.85},
number={3},
journal={MRS Communications},
author={Wang, Yiqun and Wagner, Nicholas and Rondinelli, James M.},
year={2019},
pages={793–805}
}

@article{SR-materials:2022,
title = {Symbolic regression in materials science via dimension-synchronous-computation},
journal = {Journal of Materials Science and Technology},
volume = {122},
pages = {77-83},
year = {2022},
issn = {1005-0302},
doi = {https://doi.org/10.1016/j.jmst.2021.12.052},
url = {https://www.sciencedirect.com/science/article/pii/S1005030222002055},
author = {Changxin Wang and Yan Zhang and Cheng Wen and Mingli Yang and Turab Lookman and Yanjing Su and Tong-Yi Zhang}
}

@article{SR-biology:2022,
    author = {Christensen, Niels Johan and Demharter, Samuel and Machado, Meera and Pedersen, Lykke and Salvatore, Marco and Stentoft-Hansen, Valdemar and Iglesias, Miquel Triana},
    title = {Identifying interactions in omics data for clinical biomarker discovery using symbolic regression},
    journal = {Bioinformatics},
    volume = {38},
    number = {15},
    pages = {3749-3758},
    year = {2022},
    month = {06},
    issn = {1367-4803},
    doi = {10.1093/bioinformatics/btac405},
    url = {https://doi.org/10.1093/bioinformatics/btac405},
    eprint = {https://academic.oup.com/bioinformatics/article-pdf/38/15/3749/49884306/btac405.pdf},
}

@article{SR-biology:2020,
title = "Automated Discovery of Relationships, Models, and Principles in Ecology",
author = "Pedro Cardoso and Branco, Vasco V. and Borges, Paulo A.V. and Carvalho, Jos{\'e} C. and Fran{\c c}ois Rigal and Rosalina Gabriel and Stefano Mammola and Jos{\'e} Cascalho and Lu{\'i}s Correia",
year = "2020",
month = dec,
day = "11",
doi = "10.3389/fevo.2020.530135",
language = "English",
volume = "8",
journal = "Frontiers in Ecology and Evolution",
issn = "2296-701X",
publisher = "Frontiers Media S.A.",
}

@article{SR-economics:2019,
author = {Claveria, Oscar and Monte, Enric and Torra, Salvador},
year = {2019},
month = {02},
pages = {833-849},
title = {Evolutionary Computation for Macroeconomic Forecasting},
volume = {53},
journal = {Computational Economics},
doi = {10.1007/s10614-017-9767-4}
}

@article{SR-economics:2016,
author = {Oscar Claveria and Enric Monte and Salvador Torra},
title = {Quantification of Survey Expectations by Means of Symbolic Regression via Genetic Programming to Estimate Economic Growth in Central and Eastern European Economies},
journal = {Eastern European Economics},
volume = {54},
number = {2},
pages = {171--189},
year = {2016},
publisher = {Routledge},
doi = {10.1080/00128775.2015.1136564},
URL = {https://doi.org/10.1080/00128775.2015.1136564},
eprint = {https://doi.org/10.1080/00128775.2015.1136564}
}

@book{Robust-Rousseeuw:1987,
author = {Rousseeuw, Peter and Leroy, Annick},
year = {1987},
month = {09},
pages = {},
title = {Robust Regression and Outlier Detection},
isbn = {9780471852339},
journal = {John Wiley, New York, ISBN 0-471-85233-3.},
doi = {10.2307/2289958}
}

@Inbook{Robust-Huber:1992,
author="Huber, Peter J.",
editor="Kotz, Samuel
and Johnson, Norman L.",
title="Robust Estimation of a Location Parameter",
bookTitle="Breakthroughs in Statistics: Methodology and Distribution",
year="1992",
publisher="Springer New York",
address="New York, NY",
pages="492--518",
isbn="978-1-4612-4380-9",
doi="10.1007/978-1-4612-4380-9_35",
url="https://doi.org/10.1007/978-1-4612-4380-9_35"
}

@article{Robust-Rousseeuw:1984,
author = {Rousseeuw, Peter},
year = {1984},
month = {12},
pages = {871-880},
title = {Least Median of Squares Regression},
volume = {79},
journal = {Journal of The American Statistical Association - J AMER STATIST ASSN},
doi = {10.1080/01621459.1984.10477105}
}

@article{RANSAC:1981,
author = {Fischler, Martin A. and Bolles, Robert C.},
title = {Random sample consensus: a paradigm for model fitting with applications to image analysis and automated cartography},
year = {1981},
issue_date = {June 1981},
publisher = {Association for Computing Machinery},
address = {New York, NY, USA},
volume = {24},
number = {6},
issn = {0001-0782},
url = {https://doi.org/10.1145/358669.358692},
doi = {10.1145/358669.358692},
journal = {Commun. ACM},
month = jun,
pages = {381–395},
numpages = {15}
}

@inproceedings{exRANSAC:2007,
  title={Hough-Transform and Extended RANSAC Algorithms for Automatic Detection of 3D Building Roof Planes from Lidar Data},
  author={Fayez Tarsha-Kurdi and Tania Landes and Pierre Grussenmeyer},
  year={2007},
  url={https://api.semanticscholar.org/CorpusID:893386}
}

@inproceedings{Sun:2025,
author = {Sun, Chenglu and Shen, Shuo and Tao, Wenzhi and Xue, Deyi and Zhou, Zixia},
title = {Noise-resilient symbolic regression with dynamic gating reinforcement learning},
year = {2025},
isbn = {978-1-57735-897-8},
publisher = {AAAI Press},
url = {https://doi.org/10.1609/aaai.v39i19.34280},
doi = {10.1609/aaai.v39i19.34280},
booktitle = {Proceedings of the Thirty-Ninth AAAI Conference on Artificial Intelligence and Thirty-Seventh Conference on Innovative Applications of Artificial Intelligence and Fifteenth Symposium on Educational Advances in Artificial Intelligence},
articleno = {2307},
numpages = {9},
series = {AAAI'25/IAAI'25/EAAI'25}
}

@inbook{robust-local:2019,
author = {Morgenthaler, Stephan},
year = {2019},
month = {05},
pages = {105-128},
title = {Fitting Redescending M-Estimators in Regression},
isbn = {9780203740538},
doi = {10.1201/9780203740538-5}
}

@article{robust-local:2004,
author = {Müller, Christine},
year = {2004},
month = {01},
pages = {},
title = {Redescending M-estimators in regression analysis, cluster analysis and image analysis},
volume = {24},
journal = {Discussiones Mathematicae. Probability and Statistics},
doi = {10.7151/dmps.1046}
}

@article{Keren:2023,
author = {Simon Keren, Liron and Liberzon, Alex and Lazebnik, Teddy},
year = {2023},
month = {01},
pages = {1249},
title = {A computational framework for physics-informed symbolic regression with straightforward integration of domain knowledge},
volume = {13},
journal = {Scientific Reports},
doi = {10.1038/s41598-023-28328-2}
}

@article{Zhou:2023,
author = {Zhou, Shuwei and Yang, Bing and Xiao, Shou and Guangwu, Yang and Zhu, Tao},
year = {2023},
month = {03},
pages = {},
title = {Crack Growth Rate Model Derived from Domain Knowledge-Guided Symbolic Regression},
volume = {36},
journal = {Chinese Journal of Mechanical Engineering},
doi = {10.1186/s10033-023-00876-8}
}

@article{MAP-Elites:2015,
  title={Illuminating search spaces by mapping elites},
  author={Jean-Baptiste Mouret and Jeff Clune},
  journal={ArXiv},
  year={2015},
  volume={abs/1504.04909},
  url={https://api.semanticscholar.org/CorpusID:14759751}
}

@book{Phillips:2021,
  author = {Phillips, Robert L.},
  title = {Pricing and Revenue Optimization},
  edition = {2},
  year = {2021},
  publisher = {Stanford University Press},
  isbn = {9781503610002}
}

@book{Bodea:2014,
publisher = {Routledge},
isbn = {9780415898331},
year = {2014},
title = {Segmentation, Revenue Management, and Pricing Analytics},
address = {Oxon},
author = {Bodea, Tudor},
}

@article{cmaes03,
  author  = {Hansen, Nikolaus and Müller, Sibylle D. and Koumoutsakos, Petros},
  title   = {{Reducing the Time Complexity of the Derandomized Evolution Strategy with Covariance Matrix Adaptation (CMA-ES)}},
  journal = {Evolutionary Computation},
  volume  = {11},
  number  = {1},
  pages   = {1-18},
  year    = {2003},
  doi     = {10.1162/106365603321828970}
}

@inproceedings{cmaes96,
  author    = {Hansen, N. and Ostermeier, A.},
  booktitle = {Proceedings of IEEE International Conference on Evolutionary Computation},
  title     = {Adapting arbitrary normal mutation distributions in evolution strategies: the covariance matrix adaptation},
  year      = {1996},
  series    = {ICEC '96},
  pages     = {312-317},
  doi       = {10.1109/ICEC.1996.542381}
}

@article{cully2017quality,
  title={Quality and diversity optimization: A unifying modular framework},
  author={Cully, Antoine and Demiris, Yiannis},
  journal={IEEE Transactions on Evolutionary Computation},
  volume={22},
  number={2},
  pages={245--259},
  year={2017},
  publisher={IEEE}
}

@unknown{PySR:2023,
author = {Cranmer, Miles},
year = {2023},
month = {05},
pages = {},
title = {Interpretable Machine Learning for Science with PySR and SymbolicRegression.jl},
doi = {10.48550/arXiv.2305.01582}
}

@inproceedings{parameterized:2023,
author = {Pietropolli, Gloria and Camerota Verd\`{u}, Federico Julian and Manzoni, Luca and Castelli, Mauro},
title = {Parametrizing GP Trees for Better Symbolic Regression Performance through Gradient Descent},
year = {2023},
isbn = {9798400701207},
publisher = {Association for Computing Machinery},
address = {New York, NY, USA},
url = {https://doi.org/10.1145/3583133.3590574},
doi = {10.1145/3583133.3590574},
booktitle = {Proceedings of the Companion Conference on Genetic and Evolutionary Computation},
pages = {619–622},
numpages = {4},
location = {Lisbon, Portugal},
series = {GECCO '23 Companion}
}

@inproceedings{kmeans:1967,
  title={Some methods for classification and analysis of multivariate observations},
  author={J. MacQueen},
  year={1967},
  url={https://api.semanticscholar.org/CorpusID:6278891}
}

@unknown{QD-SR:2019,
author = {Bruneton, Jean-Philippe and Cazenille, Leo and Douin, A. and Reverdy, V.},
year = {2019},
month = {06},
title = {Exploration and Exploitation in Symbolic Regression using Quality-Diversity and Evolutionary Strategies Algorithms},
doi = {10.48550/arXiv.1906.03959}
}

@unknown{QD-SR:2025,
author = {Bruneton, Jean-Philippe},
year = {2025},
month = {03},
pages = {},
title = {Enhancing Symbolic Regression with Quality-Diversity and Physics-Inspired Constraints},
doi = {10.48550/arXiv.2503.19043}
}

% \begin{comment}
\clearpage
\appendix

% ============================================================ %
\section{Effectiveness of Robust Loss Functions against Outliers for SR and MOSR} \label{sec:appendix:robust_loss}
\begin{figure*}[t]
    \centering
    \includegraphics[width=0.75\linewidth]{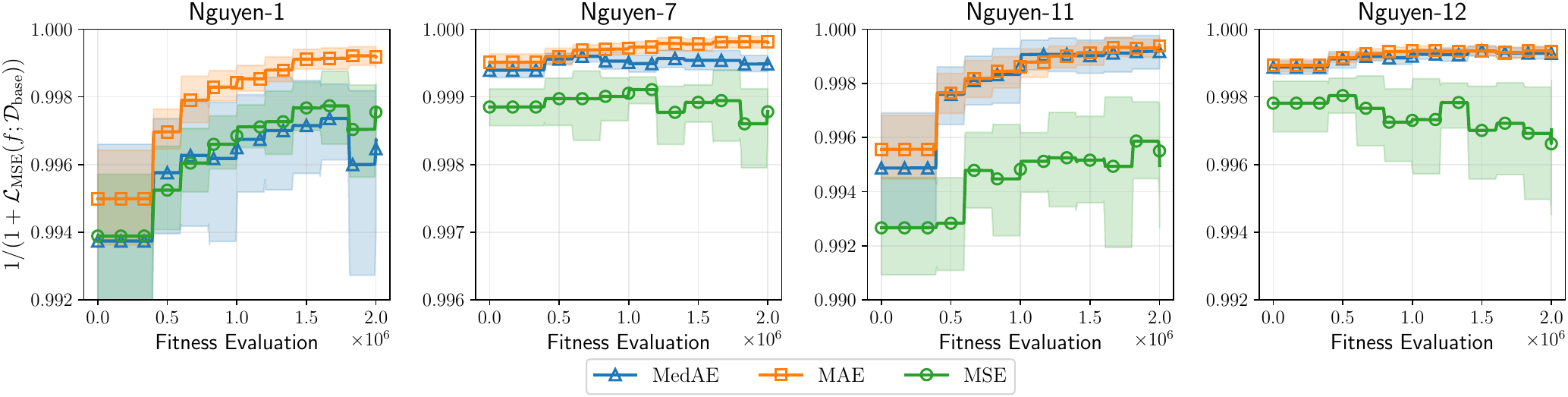}
    \caption{
        Comparison of trajectories of the best predictive accuracy in each generation of SR on Nguyen benchmarks with $n_{\mathrm{noise}} = 5$ under three loss functions: $\Loss_\mathrm{MedAE}$, $\Loss_\mathrm{MAE}$, and $\Loss_\mathrm{MSE}$.
        The solid curves show the mean accuracy, and the shaded regions show the 95\% bootstrap confidence intervals over 10 independent trials.
        The accuracy is defined as $1/(1 + \Loss_\mathrm{MSE}(f;\Data_\mathrm{base}))$.
    }
    \label{fig:nguyen_5noise:sr}
\end{figure*}

\begin{figure*}[t]
    \centering
    \includegraphics[width=0.75\linewidth]{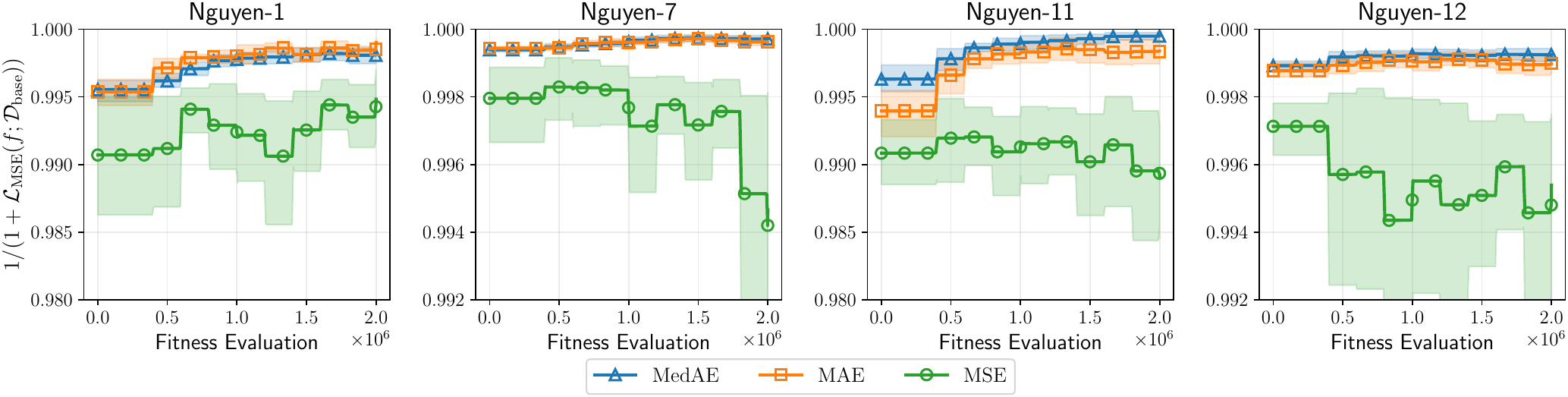}
    \caption{
        Comparison of trajectories of the best predictive accuracy in each generation of SR on Nguyen benchmarks with $n_{\mathrm{noise}} = 20$ under three loss functions: $\Loss_\mathrm{MedAE}$, $\Loss_\mathrm{MAE}$, and $\Loss_\mathrm{MSE}$.
        The solid curves show the mean accuracy, and the shaded regions show the 95\% bootstrap confidence intervals over 10 independent trials.
        The accuracy is defined as $1/(1 + \Loss_\mathrm{MSE}(f;\Data_\mathrm{base}))$.
    }
    \label{fig:nguyen_20noise:sr}
\end{figure*}

\begin{figure*}[t]
    \centering
    \includegraphics[width=0.75\linewidth]{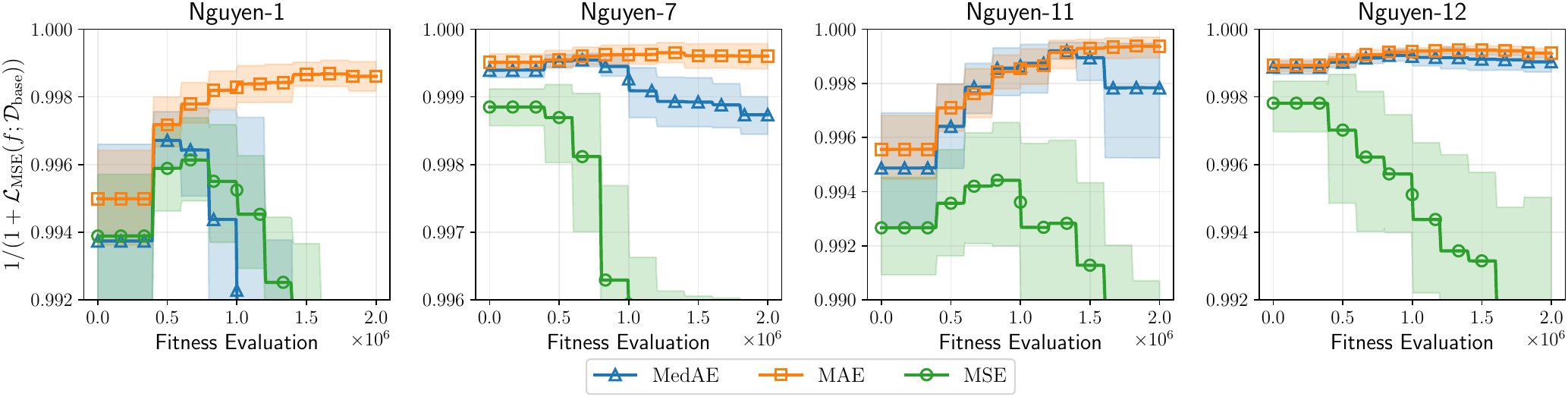}
    \caption{
        Comparison of trajectories of the best predictive accuracy in each generation of MOSR on Nguyen benchmarks with $n_{\mathrm{noise}} = 5$ under three loss functions: $\Loss_\mathrm{MedAE}$, $\Loss_\mathrm{MAE}$, and $\Loss_\mathrm{MSE}$.
        The solid curves show the mean accuracy, and the shaded regions show the 95\% bootstrap confidence intervals over 10 independent trials.
        The accuracy is defined as $1/(1 + \Loss_\mathrm{MSE}(f;\Data_\mathrm{base}))$.
    }
    \label{fig:nguyen_5noise:mosr}
\end{figure*}

\begin{figure*}[t]
    \centering
    \includegraphics[width=0.75\linewidth]{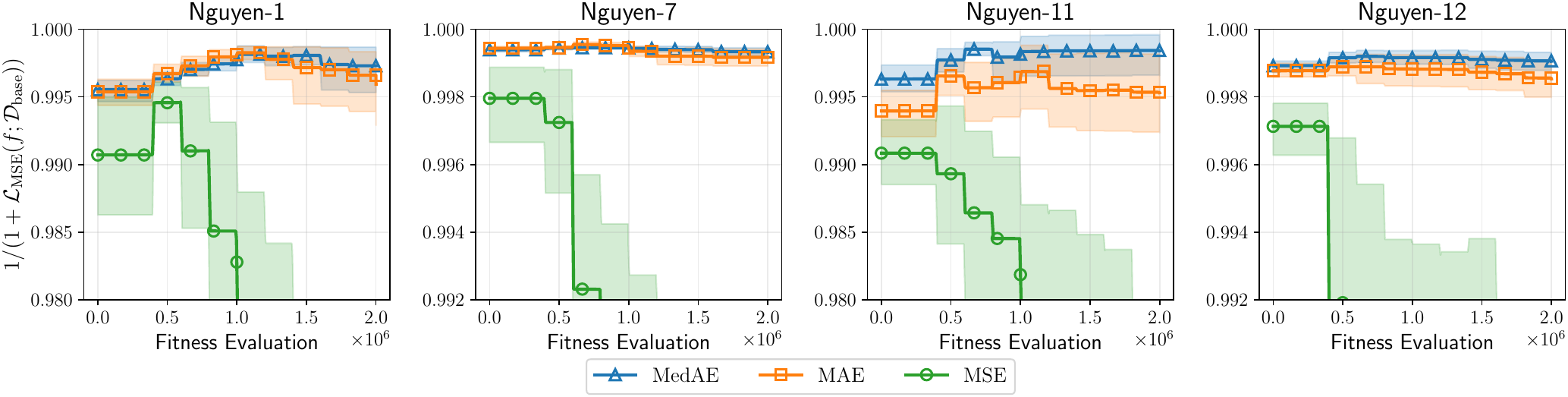}
    \caption{
        Comparison of trajectories of the best predictive accuracy in each generation of MOSR on Nguyen benchmarks with $n_{\mathrm{noise}} = 20$ under three loss functions: $\Loss_\mathrm{MedAE}$, $\Loss_\mathrm{MAE}$, and $\Loss_\mathrm{MSE}$.
        The solid curves show the mean accuracy, and the shaded regions show the 95\% bootstrap confidence intervals over 10 independent trials.
        The accuracy is defined as $1/(1 + \Loss_\mathrm{MSE}(f;\Data_\mathrm{base}))$.
    }
    \label{fig:nguyen_20noise:mosr}
\end{figure*}

% ============================================================ %
This appendix presents the corresponding SR and MOSR results for the experiment in \Cref{ssec:exp1:robust_loss}.
As shown in \Cref{fig:nguyen_5noise:sr,fig:nguyen_20noise:sr,fig:nguyen_5noise:mosr,fig:nguyen_20noise:mosr}, similar trends to those observed for DRSR can also be seen for both SR and MOSR: robust loss functions $\Loss_\mathrm{MedAE}$ and $\Loss_\mathrm{MAE}$ generally outperform $\Loss_\mathrm{MSE}$, and $\Loss_\mathrm{MedAE}$ tends to become more effective than $\Loss_\mathrm{MAE}$ as the number of outliers increases.



% ============================================================ %
\section{Representative Expressions Obtained by DRSR} \label{sec:appendix:expressions}
% ============================================================ %
\subsection{Expressions on the Mixture Dataset}
\label{ssec:appendix:expr:mixture}
This subsection reports the explicit forms of the expressions found by DRSR and shown in \Cref{fig:mixture:xy}.
They are ranked according to the component-wise accuracies on $\Data_{\mathrm{linear}}$ and $\Data_{\mathrm{logistic}}$.
For readability, numerical coefficients are rounded to three significant digits.

\paragraph{Top 3 Expressions on $\Data_{\mathrm{linear}}$}
\Cref{tab:appendix:mixture:linear} lists the three expressions found by DRSR and shown in \Cref{fig:mixture:xy} for the linear component.
The top two expressions can be simplified to forms close to the ground-truth relation $y=1-0.1x$;
for example, Rank 1 can be simplified to $y=1.00-0.0993x$, and Rank 2 can be simplified to $y=1.00-0.101x$.
\begin{table*}[tb]
    \centering
    % \small
    \caption{Explicit forms of the three expressions found by DRSR and shown in \Cref{fig:mixture:xy} for the linear component $\Data_{\mathrm{linear}}$.}
    \label{tab:appendix:mixture:linear}
    \vspace{-2mm}
    \begin{tabular}{l}
        \toprule
        $\text{Rank 1:}\ y=0.584\cdot\log(5.54\cdot\exp(-0.170x))$                                     \\
        $\text{Rank 2:}\ y=1.02\cdot(0.984+(-0.0990x))$                                                \\
        $\text{Rank 3:}\ y=0.349\cdot(2.93\cdot\exp(-2.57\cdot\exp(-4.88\cdot\exp(1.31x)))+(-0.301x))$ \\
        \bottomrule
    \end{tabular}
    \vspace{4mm}
\end{table*}
\paragraph{Top 3 Expressions on $\Data_{\mathrm{logistic}}$}
\Cref{tab:appendix:mixture:logistic} lists the three expressions found by DRSR and shown in \Cref{fig:mixture:xy} for the logistic component.
The first-ranked expression can be simplified to
\begin{align}
    y=\frac{0.988}{1+0.0176\exp(1.62x)} \enspace,
\end{align}
which has a clear logistic form and closely matches the ground-truth relation
\begin{align}
    y=\frac{1}{1+\exp(-4+1.6x)}=\frac{1}{1+0.0183\exp(1.6x)} \enspace.
\end{align}

\begin{table*}[tb]
    \centering
    \small
    \caption{Explicit forms of the three expressions found by DRSR and shown in \Cref{fig:mixture:xy} for the logistic component $\Data_{\mathrm{logistic}}$.}
    \label{tab:appendix:mixture:logistic}
    \vspace{-2mm}
    \begin{tabular}{l}
        \toprule
        $\text{Rank 1:}\ y=49.1 \cdot \dfrac{3.50}{3.43\cdot\left(50.7 + 0.891\cdot\exp(1.62x)\right)}$                                                                                                                         \\
        $\text{Rank 2:}\ y=617\cdot\dfrac{1}{1\cdot\left(1\cdot\left(652+\left(-167\cdot\dfrac{1}{1\cdot(8.23 + 0.194\cdot(x\cdot x))}\right)\right) + 1\cdot(x\cdot 1\cdot\left(0.278x + 12.4\cdot\exp(1.22x)\right))\right)}$ \\
        $\text{Rank 3:}\ y=$                                                                                                                                                                                                    \\
        $53.3\cdot\dfrac{6.96}{
                6.52\cdot\left(58.0 + 1.38\cdot\left(0.901\cdot(0.463+(-1.37\cdot(-0.330x\cdot1.88x))) \cdot 0.579\cdot\dfrac{-4.52\cdot\left( 1.18\cdot\dfrac{2.64}{5.47\cdot(-5.80+1.40x)}+(-1.61\cdot\exp(1.42x))\right)}{7.96x}\right)\right)
        }$                                                                                                                                                                                                                      \\
        \bottomrule
    \end{tabular}
    \vspace{4mm}
\end{table*}

\subsection{Expressions on the Astronomical Dataset}
\label{ssec:appendix:expressions:astronomy}
This subsection reports the explicit forms of the expressions shown in \Cref{fig:astronomy:xy,fig:astronomy:logxlogy,fig:astronomy:search_logxlogy}.
For readability, numerical coefficients are rounded to three significant digits.

\paragraph{Expressions in the Non-log-transformed Space}
\Cref{tab:appendix:astronomy} lists the six expressions shown in \Cref{fig:astronomy:xy,fig:astronomy:logxlogy}.
As discussed in \Cref{sec:exp2}, the 4th- and 6th-ranked expressions have the structure $a\exp(b\log(cx))$, corresponding to the power-law form $y=Ax^b$;
for example, Rank 4 can be simplified to $y=3.09x^{3.82}$, and Rank 6 can be simplified to $y=1.67x^{3.63}$.
\begin{table*}[tb]
    \centering
    \caption{Explicit forms of the six expressions shown in \Cref{fig:astronomy:xy,fig:astronomy:logxlogy}.}
    \label{tab:appendix:astronomy}
    \vspace{-2mm}
    \begin{tabular}{l}
        \toprule
        $\text{Rank 1:}\ y=698\cdot\exp(-20.8\cdot\exp(-4.18x))$                        \\
        $\text{Rank 2:}\ y=1.66\cdot\exp(-15.1\cdot\exp(-6.63x))$                       \\
        $\text{Rank 3:}\ y=26300\cdot(2.06x\cdot1.48\cdot\exp(-21.5\cdot\exp(-2.83x)))$ \\
        $\text{Rank 4:}\ y=1.75\cdot\exp(3.82\cdot\log(1.16x))$                         \\
        $\text{Rank 5:}\ y=2.76\cdot\exp(-13.2\cdot\exp(-6.73\cdot(4.09x\cdot2.02x)))$  \\
        $\text{Rank 6:}\ y=0.509\cdot\exp(3.63\cdot\log(1.96\cdot(0.83\cdot0.853x)))$   \\
        \bottomrule
    \end{tabular}
    \vspace{4mm}
\end{table*}

\paragraph{Expressions in the Log-transformed Space}
\Cref{tab:appendix:astronomy:log_combined} lists the six expressions shown in \Cref{fig:astronomy:search_logxlogy} together with their corresponding linear relations on the $\log M$--$\log L$ plane.
To obtain the linear relations, we invert the min--max normalization used in the log-transformed experiment, where the original variables $\log M$ and $\log L$ were normalized to $[0,1]$ as
\begin{align}
    x & =\frac{\log M+0.9682}{2.4039}, \\
    y & =\frac{\log L+2.313}{7.5}.
\end{align}
Accordingly, the inverse transformation is given by
\begin{align}
    \log M & =2.4039x - 0.9682, \\
    \log L & =7.5y - 2.313.
\end{align}
\begin{table*}[tb]
    \centering
    \caption{Explicit forms of the six expressions shown in \Cref{fig:astronomy:search_logxlogy} and the corresponding linear relations on the $\log M$--$\log L$ plane obtained by the inverse transformation.}
    \label{tab:appendix:astronomy:log_combined}
    \vspace{-2mm}
    \begin{tabular}{c l l}
        \toprule
        Rank & Expression in the normalized space & Relation on the $\log M$--$\log L$ plane \\
        \midrule
        1    & $y=1.07\cdot(-0.253+1.33x)$        & $\log L = 4.44\log M - 0.0445$           \\
        2    & $y=0.508\cdot(-0.538+2.81x)$       & $\log L = 4.45\log M - 0.0508$           \\
        3    & $y=0.691\cdot(-0.320+1.92x)$       & $\log L = 4.14\log M + 0.0362$           \\
        4    & $y=0.864\cdot(-0.101+1.26x)$       & $\log L = 3.40\log M + 0.321$            \\
        5    & $y=0.878x$                         & $\log L = 2.74\log M + 0.339$            \\
        6    & $y=0.891x$                         & $\log L = 2.78\log M + 0.378$            \\
        \bottomrule
    \end{tabular}
    \vspace{4mm}
\end{table*}

\Cref{tab:appendix:eker_mlr} summarizes the piecewise classical mass--luminosity relations reported by \citet{Eker:2018}.
Using their sample of 509 main-sequence stars, they obtained these relations by fitting six linear relations on the $\log M$--$\log L$ plane over different mass ranges.
Here, $M$ and $L$ are in solar units, and the reported coefficient errors are obtained by standard error analysis of the least-squares fit.
Comparing \Cref{tab:appendix:astronomy:log_combined} with \Cref{tab:appendix:eker_mlr}, we find that the Rank 1 and Rank 2 expressions fall within the coefficient ranges of the very-low-mass relation in \Cref{tab:appendix:eker_mlr}.
In addition, unlike Rank 1 and Rank 2, the Rank 3 expression is close to the high-mass relation:
its intercept, 0.0362, falls within the reported range $0.093 \pm 0.083$,
and its slope, 4.14, is close to the upper end of the reported range $3.967 \pm 0.143$ (i.e., 4.11).
\begin{table*}[tb]
    \centering
    \caption{Piecewise classical mass--luminosity relations reported by \citet{Eker:2018}. The relations are given on the $\log M$--$\log L$ plane, with $M$ and $L$ in solar units; the reported uncertainties are the coefficient errors.}
    \label{tab:appendix:eker_mlr}
    \vspace{-2mm}
    \begin{tabular}{l c l}
        \toprule
        Domain            & $\log M$ range      & Classical expression                                    \\
        \midrule
        Ultra low mass    & $(-0.747,\ -0.347]$ & $\log L = (2.028 \pm 0.135)\log M + (-0.976 \pm 0.070)$ \\
        Very low mass     & $(-0.347,\ -0.143]$ & $\log L = (4.572 \pm 0.319)\log M + (-0.102 \pm 0.076)$ \\
        Low mass          & $(-0.143,\ 0.0212]$ & $\log L = (5.743 \pm 0.413)\log M + (-0.007 \pm 0.026)$ \\
        Intermediate mass & $(0.0212,\ 0.380]$  & $\log L = (4.329 \pm 0.087)\log M + (0.010 \pm 0.019)$  \\
        High mass         & $(0.380,\ 0.845]$   & $\log L = (3.967 \pm 0.143)\log M + (0.093 \pm 0.083)$  \\
        Very high mass    & $(0.845,\ 1.491]$   & $\log L = (2.865 \pm 0.155)\log M + (1.105 \pm 0.176)$  \\
        \bottomrule
    \end{tabular}
\end{table*}

% \end{comment}

\end{document}